\documentclass[a4paper,fleqn]{cas-dc}

\usepackage{hyperref}
\usepackage{graphicx}
\usepackage[caption=false]{subfig}
\usepackage{lipsum}
\usepackage{lscape}
\usepackage{amsmath}
\usepackage{amssymb}
\usepackage{multirow}
\usepackage[figuresright]{rotating}
\usepackage{algorithm} 
\usepackage{algpseudocode} 
\usepackage{comment}
\usepackage{lineno}
\usepackage{nomencl}
\makenomenclature
\usepackage{calc}
\usepackage[T1]{fontenc}
\usepackage{relsize}
\usepackage{orcidlink}
\usepackage{bm}

\usepackage[sort,comma,authoryear,round]{natbib}

\setcitestyle{authoryear,open={(},close={)},citesep={;}} 

\hypersetup{
  colorlinks,
  citecolor=Violet,
  linkcolor=Red,
  urlcolor=Blue}

\newcolumntype{P}[1]{>{\centering\arraybackslash}p{#1}}

\def\tsc#1{\csdef{#1}{\textsc{\lowercase{#1}}\xspace}}
\tsc{WGM}
\tsc{QE}
\tsc{EP}
\tsc{PMS}
\tsc{BEC}
\tsc{DE}

\begin{document}
\let\WriteBookmarks\relax
\def\floatpagepagefraction{1}
\def\textpagefraction{.001}
\shorttitle{3D Gaussian Splatting for Strawberry Phenotyping}
\shortauthors{Li et al. (2025)}
 
\title [mode = title]{Object-Centric 3D Gaussian Splatting for Strawberry Plant Reconstruction and Phenotyping}


\author[1]{Jiajia Li}\ead{lijiajia@msu.edu}
\author[2]{Keyi Zhu}\ead{zhukeyi1@msu.edu}
\author[3]{Qianwen Zhang}\ead{qz72@msstate.edu}
\author[4]{Dong Chen}\ead{dc2528@msstate.edu}
\author[5]{Qi Sun}\ead{qisun@nyu.edu}
\author[2]{Zhaojian Li*}\ead{lizhaoj1@egr.msu.edu}

\address[1]{Department of Electrical and Computer Engineering, Michigan State University, East Lansing, MI, USA}
\address[2]{Department of Mechanical Engineering, Michigan State University, East Lansing, MI, USA}
\address[3]{Truck Crops Branch Experiment Station, Mississippi State University, Starkville, MS, USA}
\address[4]{Department of Agricultural and Biological Engineering, Mississippi State University, Starkville, MS, USA}
\address[5]{Tandon School of Engineering, New York University, NY, USA}

\address{* Corresponding author}

\begin{abstract}
Strawberries are among the most economically significant fruits in the United States, generating over \$2 billion in annual farm-gate sales and accounting for approximately 13\% of the total fruit production value. Plant phenotyping plays a vital role in selecting superior cultivars by characterizing plant traits such as morphology, canopy structure, and growth dynamics. However, traditional plant phenotyping methods are time-consuming, labor-intensive, and often destructive. Recently, neural rendering techniques, notably Neural Radiance Fields (NeRF) and 3D Gaussian Splatting (3DGS), have  emerged as powerful frameworks for high-fidelity 3D reconstruction. By capturing a sequence of multi-view images or videos around a target plant, these methods enable non-destructive reconstruction of complex plant architectures. Despite their promise, most current applications of 3DGS in agricultural domains reconstruct the entire scene, including background elements, which introduces noise, increases computational costs, and complicates downstream trait analysis. 
To address this limitation, we propose a novel object-centric 3D reconstruction framework incorporating a preprocessing pipeline that leverages the Segment Anything Model v2 (SAM-2) and alpha channel background masking to achieve clean strawberry plant reconstructions. This approach produces more accurate geometric representations while substantially reducing computational time. With a background-free reconstruction, our algorithm can automatically estimate important plant traits, such as plant height and canopy width, using DBSCAN clustering and Principal Component Analysis (PCA). Experimental results show that our method outperforms conventional pipelines in both accuracy and efficiency, offering a scalable and non-destructive solution for strawberry plant phenotyping.
\end{abstract}


\maketitle

\section{Introduction}
Strawberries (\textit{Fragaria} × \textit{ananassa}) are not only valued for their rich nutritional profile, being an excellent source of vitamins, minerals, and antioxidants that promote human health, but also for their strong consumer demand and economic importance \citep{giampieri2012strawberry}. Strawberries are among the most widely consumed and economically significant fruits in the United States. In 2024, U.S. strawberry production exceeded 1.6 billion pounds, with California and Florida serving as the primary production regions \citep{freshproduce2024strawberry}. However, strawberry plants and their nutritional composition are highly sensitive to environmental changes such as temperature and light intensity \citep{tulipani2011influence}.

Given the crop’s economic importance and sensitivity to environmental factors, effective cultivation and phenotyping strategies are essential to improve yield and quality. The cultivation process plays a critical role in selecting cultivars that perform best under varying environmental conditions such as temperature, humidity, and light \citep{kouloumprouka2024opportunities}. Plant phenotyping, defined as the quantitative assessment of plant traits such as morphology, physiology, and yield components, plays a critical role in cultivar development \citep{fiorani2013future}. Traditionally, these selection and evaluation processes rely heavily on manual measurements and visual assessments, which are often time-consuming, labor-intensive, and destructive \citep{liu2023molecular}. Such limitations hinder large-scale and continuous monitoring of plant growth and fruit development, underscoring the need for \textit{automated}, \textit{non-destructive}, and \textit{high-throughput} phenotyping approaches.

Recent advancements in sensing modalities, including hyperspectral imaging, LiDAR, and 3D reconstruction, combined with machine learning and deep learning algorithms, have transformed plant phenotyping into a data-rich and computationally driven discipline \citep{fiorani2013future, li2014review, jiang2020convolutional}. For instance,  \cite{ndikumana2024development} recently developed an image-based Strawberry Phenotyping Tool that integrates two deep learning architectures, YOLOv4 \citep{bochkovskiy2020yolov4} and U-Net \citep{ronneberger2015u}, into a unified system to extract multiple strawberry phenotypic traits. The system enabled the detection and measurement of six key traits, including plant height, leaf area, and flower count, either directly from natural scenes or indirectly from captured and stored images. Similarly, \cite{zheng2022deep} utilized Structure-from-Motion (SfM) techniques in combination with high-resolution RGB orthoimages, near-infrared (NIR) orthoimages, and Digital Surface Models (DSM) to enhance strawberry canopy characterization. In their study, Mask R-CNN \citep{he2017mask} was applied to orthoimages with two spectral band combinations (RGB and RGB–NIR) to accurately identify and delineate strawberry plant canopies. Despite these advances, traditional image-based and SfM approaches remain constrained by limited geometric fidelity and sensitivity to occlusion \citep{li2025survey, bao20253d}.

More recently, advanced neural rendering techniques such as Neural Radiance Fields (NeRF) and 3D Gaussian Splatting (3DGS) have emerged as powerful approaches for high-fidelity 3D reconstruction and scene representation \citep{gao2022nerf, chen2024survey, li2025survey}. 
NeRF utilizes implicit neural representations to synthesize photorealistic 3D scenes from sparse multi-view images, capturing fine geometric and textural details that conventional reconstruction methods often miss \citep{mildenhall2021nerf}. Trained in a self-supervised manner using only images and camera poses, without explicit 3D or depth annotations, NeRF is particularly advantageous for complex plant architectures where occlusion and noise hinder traditional depth-sensing approaches. 
In contrast, 3DGS models scenes as collections of Gaussian primitives, enabling efficient real-time rendering and reconstruction \citep{kerbl20233d}. By replacing volumetric rendering with point-based splatting, 3DGS achieves superior computational efficiency and scalability, making it highly suitable for high-throughput phenotyping and large-scale agricultural applications.

Although significant progress has been made, applying NeRF and 3DGS to plant phenotyping is still in its early stages and has only recently begun attracting strong interest from the agricultural research community \citep{li2025survey, jiang2025cotton3dgaussians, shen2025plantgaussian, chen2025plant}. For instance, Zhang \textit{et al.}~\citep{zhang2024neural} proposed a NeRF-based model for 3D scene reconstruction and rendering of strawberry plants. Building upon the baseline NeRF framework, their model integrates multi-resolution latent feature encoding and environmental factor embedding to enhance reconstruction quality under varying conditions. Experimental results demonstrated that the proposed NeRF model achieved photorealistic rendering performance across small-, medium-, and large-scale agricultural scenes; however, it did not specifically address plant phenotyping tasks such as trait extraction or quantitative analysis. In \cite{jiang2025cotton3dgaussians}, the authors introduced a 3DGS-based workflow for reconstructing high-fidelity 3D models of cotton plants and extracting phenotypic traits such as boll number, volume, plant height, and canopy size. Using smartphone imagery and photogrammetry, but the background was removed manually. The method achieved superior rendering quality and accurate trait estimation, with errors under 10\% compared to LiDAR ground truth. Despite these advancements, strawberry plant phenotyping remains largely unexplored using 3DGS techniques. Moreover, most current applications in the agricultural domain focus on reconstructing entire scenes, including background elements, which introduces noise, increases computational costs, and complicates downstream trait analysis.

In this paper, we propose a novel high-throughput strawberry plant phenotyping framework based on 3D Gaussian Splatting (3DGS). Instead of reconstructing the entire captured scene, our method introduces an object-centric 3D reconstruction framework that focuses on generating clean and accurate strawberry plant models from noisy backgrounds. A preprocessing pipeline leveraging the Segment Anything Model v2 (SAM-2) \citep{ravi2024sam} is incorporated prior to reconstruction to isolate plant regions. During the reconstruction, RGBA-based loss masking, opacity-guided Gaussian culling \citep{nerfstudio}, and background randomization are employed to further suppress background artifacts. This framework yields high-fidelity, object-centric strawberry plant reconstructions with improved accuracy and substantially reduced computational time.
By reconstructing background-free plant models, the proposed framework can automatically estimate key phenotypic traits, such as plant height and canopy width, using Density-Based Spatial Clustering of Applications with Noise (DBSCAN) \citep{ester1996density} and Principal Component Analysis (PCA) \citep{mackiewicz1993principal}. The framework provides a low-cost, automated, and scalable solution for precise strawberry plant analysis, paving the way for advanced 3D phenotyping and intelligent crop breeding, with potential applicability to other crop species.

\section{Materials and Methods}
\label{sec2}
\subsection{Data acquisition}
All strawberry plant data used in this study were collected under controlled indoor conditions to ensure consistent illumination and minimize environmental variability. The plants were maintained in their natural growth state without pruning, defoliation, or any other human intervention, thereby preserving their authentic canopy structures for realistic 3D reconstruction and phenotyping analyses.
Each strawberry plant was placed individually on a circular bin to provide uniform background separation and facilitate complete multi-view capture. A 10 cm calibration cube, with 9.6 cm ArUco markers affixed to all six faces, was positioned adjacent to each plant to serve as a geometric scale reference for metric restoration and alignment during 3D reconstruction.

Video data were acquired using an Apple iPhone 16. The recording resolution was set to 2160 $\times$ 3840 pixels (4K), with a frame rate of 24 frames per second (fps) to capture fine spatial and temporal details. During acquisition, the operator circumnavigated each plant along a smooth trajectory at three distinct height levels (i.e., low (approximately 0–5 cm above the soil), mid (5–20 cm), and high (20–50 cm), to ensure adequate multi-view coverage of the canopy, fruiting zones, and crown region. The data collection was conducted in April 2025. In total, 15 healthy strawberry plants at the fruiting growth stage were recorded. The resulting dataset encompasses a wide range of morphological variations, including differences in leaf density, occlusion patterns, and fruit positioning, thereby providing a robust foundation for evaluating 3D reconstruction accuracy, scale calibration, and subsequent phenotypic analysis.


\begin{figure*}[!ht]
  \centering
  \includegraphics[width=0.95\textwidth]{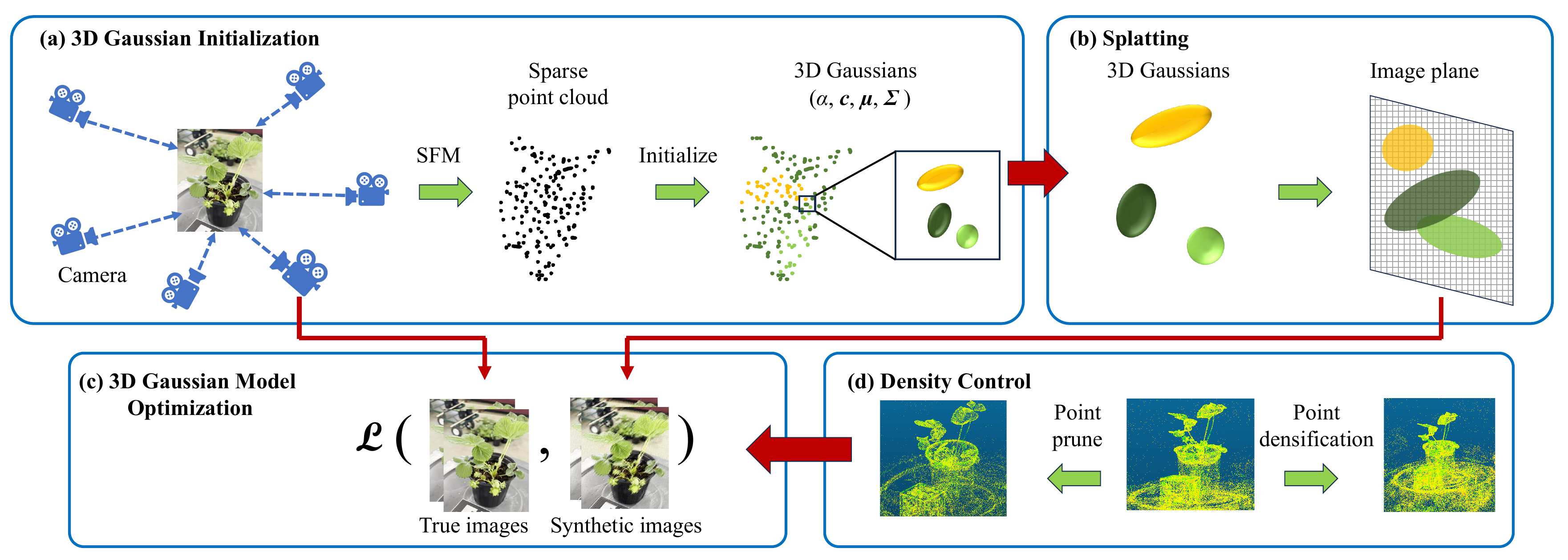}
  \vspace{1pt}
  \caption{Framework of 3D Gaussian Splatting (3DGS) \citep{kerbl20233d}. The process consists of four main stages: (a) 3D Gaussian initialization: multi-view images are used to generate a sparse point cloud via structure-from-motion (SfM), from which 3D Gaussians are initialized with parameters $(\alpha, \bm{c}, \bm{\mu}, \bm{\Sigma})$; (b) Splatting: the Gaussians are projected onto the image plane for differentiable rendering; (c) 3D Gaussian model optimization: the parameters are iteratively optimized by minimizing the discrepancy between rendered and ground-truth images; and (d) Density control: point pruning and densification maintain efficient and accurate scene representation.}
  \label{fig:3dgs}
\end{figure*}
\subsection{3D Gaussian Splatting (3DGS)}
NeRF-based methods have demonstrated remarkable 3D reconstruction fidelity by learning a volumetric scene function that maps spatial coordinates and viewing directions $(\bm{x}, \bm{d})$ to color and density $(\bm{c}, \sigma)$, without explicitly modeling scene geometry \citep{mildenhall2021nerf}. Despite their success, these approaches often suffer from heavy computational costs and long training and rendering times \citep{barron2021mip, gao2022nerf, li2025survey}. 

To overcome these limitations, 3DGS \citep{kerbl20233d} introduces an explicit 3D geometry representation that enables real-time rendering, compact scene modeling, and high reconstruction accuracy. Unlike implicit radiance field methods, 3DGS formulates an explicit radiance field by directly associating radiance information with points in 3D space (Fig.~\ref{fig:3dgs}). The model is trained in a supervised manner to minimize discrepancies between synthesized and ground-truth multi-view images, substantially improving learning efficiency and rendering speed for complex, high-resolution scenes.

As illustrated in Figure~\ref{fig:3dgs}(a), the scene geometry is represented as a set of 3D Gaussians (ellipsoids), each defined by radiance attributes, opacity $(\alpha)$ and color $(\bm{c})$, and spatial attributes, including the center $(\bm{\mu} = (\mu_x, \mu_y, \mu_z))$ and a 3D covariance matrix $(\bm{\Sigma})$. In contrast to NeRF, which infers opacity implicitly from volumetric density $(\sigma)$, 3DGS models the opacity $(\alpha)$ explicitly as a learnable parameter, allowing finer control over transparency and surface boundaries. The color component $(\bm{c})$ is represented through spherical harmonic coefficients to capture view-dependent appearance variations efficiently. To guarantee that the covariance matrix $\bm{\Sigma}$ remains positive semi-definite and geometrically meaningful, it is factorized as:
\begin{equation}
\bm{\Sigma} = \bm{R}\bm{S}\bm{S}^T\bm{R}^T,
\end{equation}
where $\bm{R}$ denotes a rotation matrix (commonly parameterized by a quaternion), and $\bm{S}$ is a diagonal scaling matrix.

In the rendering stage, splatting is employed to project each 3D Gaussian onto the 2D image plane, as illustrated in Figure~\ref{fig:3dgs}(b). To reduce redundant computation, frustum culling \citep{assarsson2000optimized} efficiently filters out Gaussians that fall outside the camera’s viewing frustum. This rendering strategy stands in contrast to NeRF’s ray-marching procedure, which requires dense sampling along camera rays and thus incurs higher computational overhead. After projection, every 3D Gaussian corresponds to a 2D elliptical Gaussian on the image plane. The final pixel color is determined through alpha compositing, where the contributions of overlapping Gaussians are blended according to their opacities \citep{zheng2024gps}:
\begin{equation}
\bm{C} = \sum_{i=1}^{|\mathcal{S}|} \bm{c}_i \alpha'_i \prod_{j=1}^{i-1}\big(1-\alpha'_j\big),
\end{equation}
with the pixel-level opacity $\alpha'_i$ computed as
\begin{equation}
    \alpha'_i = \alpha_i \cdot e^{-\frac{1}{2}(\bm{x}'-\bm{\mu}'_i)^T \bm{\Sigma}_i^{'-1}(\bm{x}'-\bm{\mu}'_i)}.
\end{equation}
This formulation allows 3DGS to achieve fast, differentiable rendering while maintaining photorealistic image quality.

To achieve real-time rendering, 3DGS leverages tile-based rasterization \citep{lassner2021pulsar} in conjunction with highly parallel CUDA-based rendering \citep{kerbl20233d}. During model optimization, density control strategies \citep{rota2024revising}, including Gaussian densification and pruning, dynamically adjust point distributions according to gradient magnitude and opacity to maintain both fidelity and efficiency (Figure~\ref{fig:3dgs}(d)). Given the ground truth RGB image $\mathbf{C}$ and rendered RGB image $\hat{\mathbf{C}}$, the overall objective function typically integrates an $\mathcal{L}_1$ color reconstruction term with a D-SSIM perceptual loss to balance pixel accuracy and structural similarity (Figure~\ref{fig:3dgs}(c)):
\begin{equation}
\begin{aligned}
\mathcal{L} 
&= (1-\lambda)\mathcal{L}_1 + \lambda \mathcal{L}_{D\text{-}SSIM} \\[4pt]
&= (1-\lambda)\lVert \mathbf{C} - \hat{\mathbf{C}} \rVert_1  
   + \lambda \big(1 - \text{SSIM}(\mathbf{C} , \hat{\mathbf{C}})\big).
\end{aligned}
\end{equation}
The training process generally follows three key stages: (i) acquiring multi-view images and initializing Gaussians (through SfM or random seeding), (ii) projecting Gaussians to synthesize novel views, and (iii) iteratively optimizing Gaussian parameters while refining their spatial density based on learned gradients and opacity cues.

\subsection{Object-centric 3DGS}\label{sec:oc_3dgs}
For the classical 3DGS, both foreground and background objects are reconstructed simultaneously within the same scene representation. While this holistic modeling captures complete environmental context, it can introduce redundant Gaussian primitives in background regions that are irrelevant to the primary object of interest. Such redundancy increases memory usage, slows optimization, and may degrade rendering quality due to unnecessary occlusions or scattering effects \citep{markin2024t, rogge2025object, jain2024gaussiancut}. In the context of strawberry plant reconstruction and phenotyping, reconstructing the entire scene is typically unnecessary and may even be detrimental, as background clutter introduces visual noise around the plant canopy. This noise complicates accurate morphological trait estimation, particularly for fine structures such as leaves, crowns, and fruit surfaces.

To overcome these limitations, we develop an object-centric 3DGS framework tailored for precise plant reconstruction. Multi-view images are preprocessed using SAM 2 \citep{ravi2024sam} to generate RGBA inputs, where the alpha channel encodes a binary mask that distinguishes the plant and reference cube (foreground) from the surrounding environment (background). Unlike the original 3DGS implementation \citep{kerbl20233d}, which relies solely on RGB supervision across all pixels, our proposed model incorporates the alpha channel into its training pipeline as a foreground supervision mask. This mechanism ensures that optimization focuses exclusively on object regions while automatically suppressing background Gaussians through the refinement process.

Given a ground-truth RGBA image $ [\mathbf{C}, \alpha_{img}]$, where $\mathbf{C}$ denotes the RGB channels and $\alpha_{img}$ represents the alpha mask channel. In addition, a rendered RGB prediction is denoted by $\hat{\mathbf{C}}$. 
our proposed model applies the $\alpha_{img}$ channel as a multiplicative binary mask during loss computation:
\begin{equation}
\begin{aligned}
\mathcal{L} 
&= \alpha_{img} \Big[(1-\lambda)\lVert \mathbf{C} - \hat{\mathbf{C}} \rVert_1  
   + \lambda \big(1 - \text{SSIM}(\mathbf{C} , \hat{\mathbf{C}})\big)
   \Big].
\end{aligned}
\end{equation}

Pixels with $\alpha_{img} = 0$ (background) are excluded from the loss, ensuring that gradients are propagated only through the foreground regions. Consequently, Gaussians projected onto background pixels receive negligible gradient updates and are not optimized during training.

An adaptive refinement strategy \citep{nerfstudio} is further employed, where it
periodically removes or splits Gaussians based on their learned opacity and gradient magnitudes. Although the RGBA alpha mask $\alpha_{img}$ is not explicitly used in this stage, its influence propagates indirectly through the optimization gradients: Gaussians corresponding to background pixels (masked by $\alpha_{img}=0$) accumulate low opacity values and are automatically pruned during the refinement step when
\begin{equation}
    \text{opacity}_i < \tau_{\alpha},
    \quad \text{with typical threshold } \tau_{\alpha} = 0.1.
\end{equation}
This synergy between masked supervision and opacity-based pruning leads to a natural elimination of background Gaussians without requiring explicit segmentation or post-processing.

During each iteration, a random background color $\mathbf{B}$ (with normalized RGB values) is used to further prevent the model from fitting to static background regions:
\begin{equation}
    \hat{\mathbf{C}} = \mathbf{C}_{\text{splat}} + (1 - \alpha_{\text{acc}}) \mathbf{B},
\end{equation}
where $\alpha_{\text{acc}}(p)$ denotes the accumulated Gaussian opacity. $\mathbf{C}_{\text{splat}}$ represents the rendered color from Gaussian splatting and $\hat{\mathbf{C}}$ denotes the final composited RGB image.
Randomizing $\mathbf{B}$ discourages the network from reconstructing the background, thereby reinforcing the suppression of non-object Gaussians.

Through the combination of (i) loss masking using RGBA alpha channels, (ii) opacity-guided Gaussian culling, and (iii) background randomization, the proposed approach achieves clean, object-centric reconstructions.  In contrast, traditional 3DGS pipelines lacking alpha masking optimize over all image pixels and thus tend to reconstruct unwanted background regions.  Consequently, the reconstructed scenes exhibit a clean separation between plant structures and background, yielding compact, geometrically accurate, and visually consistent point clouds, well-suited for downstream strawberry phenotyping and structural trait analysis.

\subsection{Plant trait estimation}
After 3D reconstruction using the proposed object-centric 3DGS, a high-fidelity point cloud of the strawberry plant and reference cube is generated. This point cloud serves as the foundation for quantitative trait estimation, including plant height and crown width.

To begin, the exported point cloud is segmented using the Density-Based Spatial Clustering of Applications with Noise (DBSCAN) algorithm \citep{ester1996density} to separate the plant and reference cube. DBSCAN is particularly suitable for this task because it does not require a predefined number of clusters and effectively handles outliers caused by minor reconstruction artifacts. Each identified cluster is analyzed by its oriented bounding box (OBB) dimensions, from which the cube cluster is isolated based on its characteristic geometric regularity and compactness.

Once the cube cluster is identified, it is used to perform scale calibration, converting the reconstruction units (arbitrary 3DGS coordinates) into physical dimensions (centimeters). Specifically, we estimate the cube’s edge length from the reconstructed geometry and compare it with the known real-world cube edge length of $10~\mathrm{cm}$. To obtain an accurate edge estimate, a plane-based measurement method is employed: multiple parallel planes are fitted to the cube’s faces, and the mean inter-plane distances are used to compute the average reconstructed edge length. This measured edge length, denoted as $\hat{l}_{\mathrm{cube}}$, yields a global scaling ratio:
\begin{equation}
s = \frac{10}{\hat{l}_{\mathrm{cube}}}.
\end{equation}
All 3D points are then rescaled by this factor $s$, ensuring that the reconstructed plant dimensions reflect true metric measurements.

Following scale restoration, the plant cluster is analyzed to extract morphological traits. The height of the plant is computed as the difference between the maximum and minimum $z$-coordinates of the scaled point cloud, while the crown width is estimated by projecting the points onto the ground plane and measuring the maximum lateral extent along principal axes obtained via Principal Component Analysis (PCA) \citep{abdi2010principal}. Additional derived metrics, such as canopy volume, leaf spread area, and fruit distribution density, can also be calculated depending on the desired phenotyping objectives.

This integrated pipeline, combining 3DGS-based reconstruction, DBSCAN segmentation, plane-based scaling, and geometric feature extraction, enables robust, accurate, and fully automated quantification of plant morphological traits. It provides a reliable foundation for downstream analyses in high-throughput phenotyping, growth monitoring, and genotype-to-phenotype association studies.

\subsection{Evaluation metrics}
Comprehensive evaluation of both 3D reconstruction quality and plant trait measurement accuracy is important for assessing the effectiveness of reconstruction frameworks in plant phenotyping. The first group of metrics focuses on 3D reconstruction evaluation, measuring how accurately the reconstructed renderings represent real-world scenes in terms of visual and structural fidelity. The second group centers on plant trait estimation, quantifying the accuracy of morphological traits (i.e., plant height and width) derived from the reconstructed 3D models. Together, these complementary metrics provide a holistic assessment of both the visual realism and biological reliability of the reconstruction pipeline.

\subsubsection{3D Reconstruction metrics}
\textit{Peak Signal-to-Noise Ratio (PSNR):}  
PSNR is a standard quantitative metric derived from the mean squared error (MSE) that expresses the fidelity of reconstructed images on a logarithmic scale. A higher PSNR indicates that the reconstruction retains more image detail and introduces less distortion or noise:
\begin{equation}
\text{PSNR} = 10 \cdot \log_{10}\!\Bigl(\frac{\text{MAX}_I^2}{\text{MSE}}\Bigr),
\end{equation}
where \(\text{MAX}_I\) represents the maximum possible pixel intensity. In the context of 3D plant reconstruction, PSNR evaluates how faithfully the generated textures and surfaces replicate the reference imagery \citep{zhao2024exploring}.

\textit{Structural Similarity Index Measure (SSIM):}
SSIM quantifies perceptual similarity by jointly considering luminance, contrast, and structural information between the reconstructed and reference images:
\begin{equation}
\text{SSIM}(x,y) = \frac{(2\mu_x\mu_y + C_1)(2\sigma_{xy} + C_2)}{(\mu_x^2+\mu_y^2+C_1)(\sigma_x^2+\sigma_y^2+C_2)},
\end{equation}
where \(\mu_x, \mu_y\) denote mean intensities, \(\sigma_x^2, \sigma_y^2\) are variances, and \(\sigma_{xy}\) represents covariance between images \(x\) and \(y\). Higher SSIM values correspond to improved perceptual quality and structural consistency \citep{zhao2024exploring}. 

\textit{Learned Perceptual Image Patch Similarity (LPIPS):}  
LPIPS measures perceptual similarity using deep feature representations extracted from pretrained convolutional neural networks. Unlike PSNR or SSIM, which operate on pixel intensity, LPIPS captures high-level semantic differences between images:
\begin{equation}
\text{LPIPS}(x,y)=\sum_{l}\frac{1}{H_lW_l}\sum_{h,w}\|w_l \odot(\hat{x}_{hw}^{l}-\hat{y}_{hw}^{l})\|_2^2,
\end{equation}
where \(\hat{x}_{hw}^{l}\) and \(\hat{y}_{hw}^{l}\) are normalized feature maps from layer \(l\) of a pretrained network, and \(w_l\) are learned channel-wise weights. Lower LPIPS scores indicate closer perceptual alignment and greater visual realism \citep{chopra2024agrinerf}.

\subsubsection{Plant trait-specific metrics}
\textit{Coefficient of Determination (\(R^2\)):}  
\(R^2\) measures the proportion of variance in observed trait values (e.g., plant height, leaf area) that is explained by predictions derived from the reconstructed model:
\begin{equation}
R^2 = 1 - \frac{\sum_{i=1}^{N}(y_i - \hat{y}_i)^2}{\sum_{i=1}^{N}(y_i - \bar{y})^2},
\end{equation}
where $N$ is the number of test samples, $\hat{y}_i$ and $y_i$ represent the estimated plant height/width and the actual plant height/width in the $i$th test image, respectively. Values approaching ``1'' indicate high predictive accuracy and strong agreement between reconstructed and ground-truth traits \citep{zhu2024three, yang2024paniclenerf}.

\textit{Root Mean Squared Error (RMSE):}  
RMSE quantifies the average magnitude of prediction error, reflecting the precision of quantitative trait estimation:
\begin{equation}
\text{RMSE} =\sqrt{\frac{1}{N}\sum_{i=1}^{N}\bigl({y_i - \hat{y}_i}\bigr)^2.}
\end{equation}
Lower RMSE values imply more accurate estimations and are particularly important for applications such as biomass prediction and morphological assessment \citep{yang2024paniclenerf}.

\textit{Mean Absolute Percentage Error (MAPE):}  
MAPE measures the average percentage deviation between predicted and observed values, providing an interpretable indicator of relative prediction error:
\begin{equation}
\text{MAPE} = \frac{1}{N}\sum_{i=1}^{N}\left|\frac{y_i - \hat{y}_i}{y_i}\right| \times 100\%.
\end{equation}
A lower MAPE value reflects higher predictive robustness and reliability of trait estimation \citep{choi2024nerf}.

\textit{Mean Absolute Error (MAE):} 
MAE measures the accuracy of the predicted
plant height or width across the test dataset, indicating how close the predictions are to the ground truth counts.
\begin{equation}
\text{MAE} = \frac{1}{N} \sum_{i=1}^{N} \left| y_i - \hat{y}_i \right|.
\end{equation}
A lower value indicates higher accuracy of the trait estimation. 

\textit{Accuracy (Acc):} accuracy evaluates how well a predictive model aligns with the actual values.
\begin{equation}
    \mathrm{Acc} = 1 - \mathrm{MAPE} 
    = 1 - \frac{1}{N}\sum_{i=1}^{N}\left|\frac{y_i - \hat{y}_i}{y_i}\right| \times 100\%.
    \label{eq:acc}
\end{equation}

\subsection{Experimental setups}
The collected strawberry videos were first decomposed into image frames, yielding approximately 100–150 frames per sequence to ensure sufficient multi-view coverage of each plant. During video acquisition, the camera was moved smoothly around each plant at multiple height levels and viewing angles to capture comprehensive geometric and texture information.
The camera intrinsics and extrinsics were estimated using Structure-from-Motion (SfM) implemented in the COLMAP pipeline, producing accurate camera poses and a sparse 3D point cloud for Gaussian initialization. The resulting image sets were subsequently divided into training (60\%) and testing (40\%) subsets. The training subset was used for model reconstruction, whereas the testing subset served to evaluate the rendering and trait estimation performance of the reconstructed 3D model.

We adopt the Splatfacto model in Nerfstudio~\citep{nerfstudio} as the implementation of the 3D Gaussian Splatting (3DGS) framework. To evaluate the effectiveness of our object-centric approach, we also implement a post background removal method (Splatfacto-PBR), in which backgrounds are removed using the SAM-2 model \textit{after} reconstruction, whereas in our method, background removal is performed \textit{prior to} reconstruction.

To establish comparative baselines, we implement three NeRF-based models: Nerfacto~\citep{nerfstudio}, Instant-NGP~\citep{muller2022instant}, and Mip-NeRF~\citep{barron2021mip}. Specifically, Nerfacto, developed within the Nerfstudio framework~\citep{nerfstudio}, integrates several state-of-the-art NeRF techniques such as hash encoding, proposal sampling, and per-image appearance conditioning to balance reconstruction quality, training efficiency, and memory usage. Instant-NGP~\citep{muller2022instant} accelerates NeRF training and inference through multi-resolution hash encoding, enabling a compact architecture that achieves high-quality results with significantly reduced computational cost. Mip-NeRF~\citep{barron2021mip} extends the NeRF framework to a continuous-scale representation by rendering anti-aliased conical frustums instead of discrete rays, effectively mitigating aliasing artifacts and improving the reconstruction of fine structural details.

All models were trained on the same strawberry dataset for 30,000 iterations, using identical training and testing splits. The Adam optimizer was used throughout, and input images were downsampled to one-quarter of their original resolution to enhance memory efficiency. The reconstruction experiments were conducted on a Linux workstation equipped with an NVIDIA GeForce RTX 5090 GPU (32 GB VRAM), Intel Xeon W-2295 CPU, and 128 GB RAM. The environment utilized CUDA 12.9 and PyTorch 2.4, providing optimized support for GPU-based parallel rendering.

\section{Results}
\label{sec3}
In this section, we present and analyze the results of the proposed framework. First, we compare the reconstruction performance of the Splatfacto and NeRF baselines to demonstrate the advantages of the 3DGS approach in terms of rendering quality and efficiency. Next, we evaluate the impact of different background removal strategies on reconstruction accuracy and scene cleanliness. Finally, we examine the effectiveness of the reconstructed 3D models in plant trait estimation.

\subsection{Comparison between 3DGS and NeRFs}
We first compare the 3DGS model, implemented via Splatfacto, with three representative NeRF-based approaches, Nerfacto, Instant-NGP, and Mip-NeRF, using the strawberry dataset. The evaluation results, averaged over 15 individual strawberry plants, are summarized in Table~\ref{tab:benchmark}.

Overall, Splatfacto demonstrates superior reconstruction performance across all key metrics. It achieves the highest PSNR (28.69) and SSIM (0.9249) values, indicating greater photometric accuracy and structural fidelity, while maintaining the lowest LPIPS (0.0781) score, reflecting strong perceptual similarity with ground-truth images. In addition, Splatfacto offers the fastest training time (8m52s) and highest rendering speed (2.83 FPS) among all evaluated models, while consuming the least GPU memory (1.89 GB).
These results highlight the computational efficiency and reconstruction quality of 3DGS compared with traditional NeRF-based approaches. The explicit Gaussian representation not only accelerates training and rendering but also enables high-fidelity 3D reconstructions suitable for detailed plant structure analysis and phenotyping.

\begin{figure*}[!ht]
  \centering
  \includegraphics[width=0.95\textwidth]{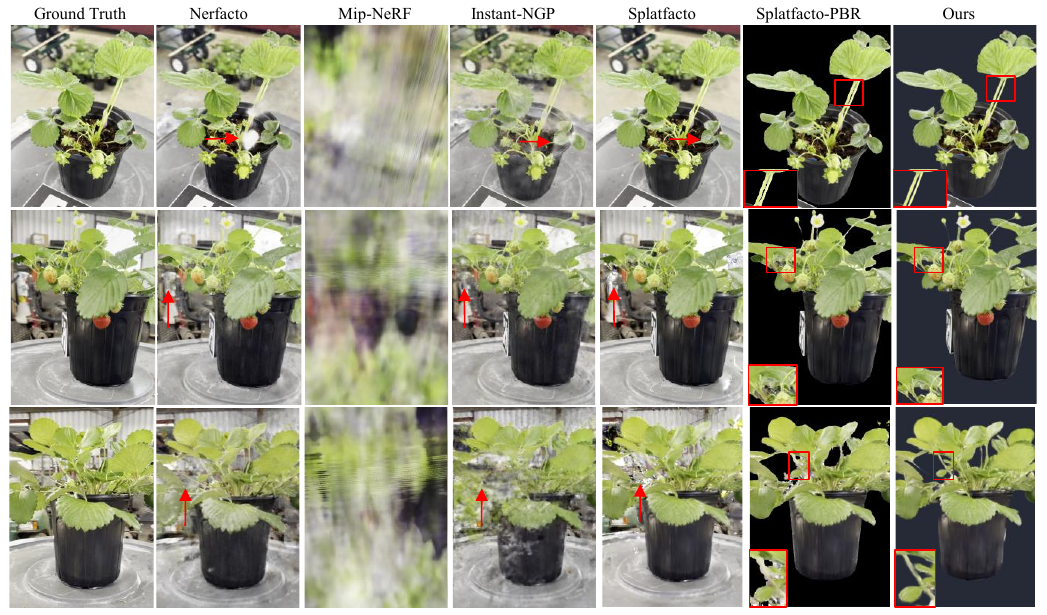}
  \vspace{1pt}
  \caption{Qualitative comparison of reconstructed strawberry plants using different NeRF-based and 3DGS-based models. From left to right: Ground Truth, Nerfacto, Mip-NeRF, Instant-NGP, Splatfacto, Splatfacto with post background removal (Splatfacto-PBR), and our proposed object-centric 3DGS. The red arrows highlight blurred or missing regions in NeRF-based reconstructions, while the red boxes emphasize finer structural details, such as leaf edges, petioles, and fruit surfaces, accurately preserved by our method.}
  \label{fig:benchmark}
\end{figure*}

Figure~\ref{fig:benchmark} illustrates qualitative reconstruction results for representative strawberry plants across various NeRF-based and 3DGS-based models. As shown, Mip-NeRF fails to preserve geometric or textural fidelity, producing severe blurring and aliasing artifacts. Instant-NGP and Nerfacto improve visual quality but still exhibit noticeable background noise and ghosting effects around leaf edges and stems (indicated by red arrows). In contrast, Splatfacto yields substantially sharper reconstructions with finer structural detail and more realistic shading, consistent with its superior quantitative performance in Table \ref{tab:benchmark}.

\begin{table*}[h]
\renewcommand{\arraystretch}{1.6}
\centering
\caption{Comparison of 3DGS performance with other NeRF models. The upward arrow (↑) indicates that higher values represent better performance, while the downward arrow (↓) denotes that lower values are preferred.}
\label{tab:benchmark}
\resizebox{0.85 \textwidth}{!}{%
\begin{tabular}{l|ccc|ccc}
\hline
            & \textbf{PSNR(↑)} & \textbf{SSIM(↑)} & \textbf{LPIPS(↓)} & \textbf{Training Time (↓)} & \textbf{Inference Rendering (FPS ↑)} & \textbf{GPU Memory (GB ↓)} \\ \hline \hline
Nerfacto    & 21.53            & 0.7738           & 0.2103            & 14m46s         & 0.7760       & 2.04         \\ 
Instant-NGP & 20.25            & 0.7890           & 0.2780            & 14m37s         & 0.7413       & 2.69         \\ 
Mip-Nerf    & 10.98            & 0.5487           & 0.8406            & 1h24m          & 0.0516       & 4.69         \\ \hline 
Splatfacto  & \textbf{28.69}            & \textbf{0.9249 }          & \textbf{0.0781}            & \textbf{8m52s}          & \textbf{2.8310}       & \textbf{1.89}         \\ \hline
\end{tabular}
}
\end{table*}

\subsection{Comparison between different background removal methods}
To further evaluate the effectiveness of the proposed object-centric 3DGS, we compared it with a post-processing background removal approach based on the original Splatfacto framework. In the post-processing setup, background pixels were removed after model training using an external pre-trained model SAM 2, while our method integrates the background suppression directly into the training pipeline as discussed in Sec.~\ref {sec:oc_3dgs}.

As shown in Table~\ref{tab:inference_time}, our method consistently outperforms the post-processing background removal across all evaluation metrics. The PSNR and SSIM scores increase by 4.19 dB and 0.0132, respectively, indicating higher reconstruction fidelity and better structural consistency. The LPIPS value decreases from 0.0657 to 0.0550, confirming improved perceptual quality. Moreover, our integrated masking approach achieves a 40\% faster training time (6m26s vs. 8m52s) and a 2.5× increase in rendering speed (7.08 FPS vs. 2.83 FPS), while maintaining comparable memory usage. 
The superior performance can be attributed to the early incorporation of background masks during optimization, which eliminates irrelevant gradients from background regions and allows the model to focus computation on the foreground plant geometry. This integrated strategy not only enhances visual quality but also improves efficiency by reducing the number of redundant Gaussians generated in non-informative regions.

\begin{table*}[h]
\renewcommand{\arraystretch}{1.6}
\centering
\caption{Comparison of our proposed background removal method with Splatfacto post-processing background removal (Splatfacto-PBR).}
\label{tab:inference_time}
\resizebox{0.85 \textwidth}{!}{%
\begin{tabular}{l|ccc|ccc}
\hline
                               & \textbf{PSNR(↑)} & \textbf{SSIM(↑)} & \textbf{LPIPS(↓)} & \textbf{Training Time (↓)} & \textbf{Rendering Speed (FPS, ↑)} & \textbf{GPU Memory (GB, ↓)} \\ \hline \hline
Splatfacto-PBR   & 23.20             & 0.9289           & 0.0657            & 8m52s          & 2.8310       & 1.89         \\ 
Ours & \textbf{27.39}            & \textbf{0.9421}           & \textbf{0.055}            & \textbf{6m26s}          & \textbf{7.0824}       & \textbf{1.96}         \\ \hline
\end{tabular}
}
\end{table*}

Furthermore, as shown in Figure~\ref{fig:benchmark}, incorporating background removal (Splatfacto-PBR) results in clearer object boundaries and reduced clutter, demonstrating the benefits of masking out irrelevant regions after training. Our proposed object-centric 3DGS (``Ours'') produces the cleanest and most photorealistic renderings, effectively isolating the plant canopy and fruit structures from the background. The red insets highlight that our method accurately preserves small details such as leaf serrations, petiole curvature, and fruit surfaces, which are essential for precise morphological analysis.

\begin{figure*}[!ht]
  \centering
  \includegraphics[width=0.95\textwidth]{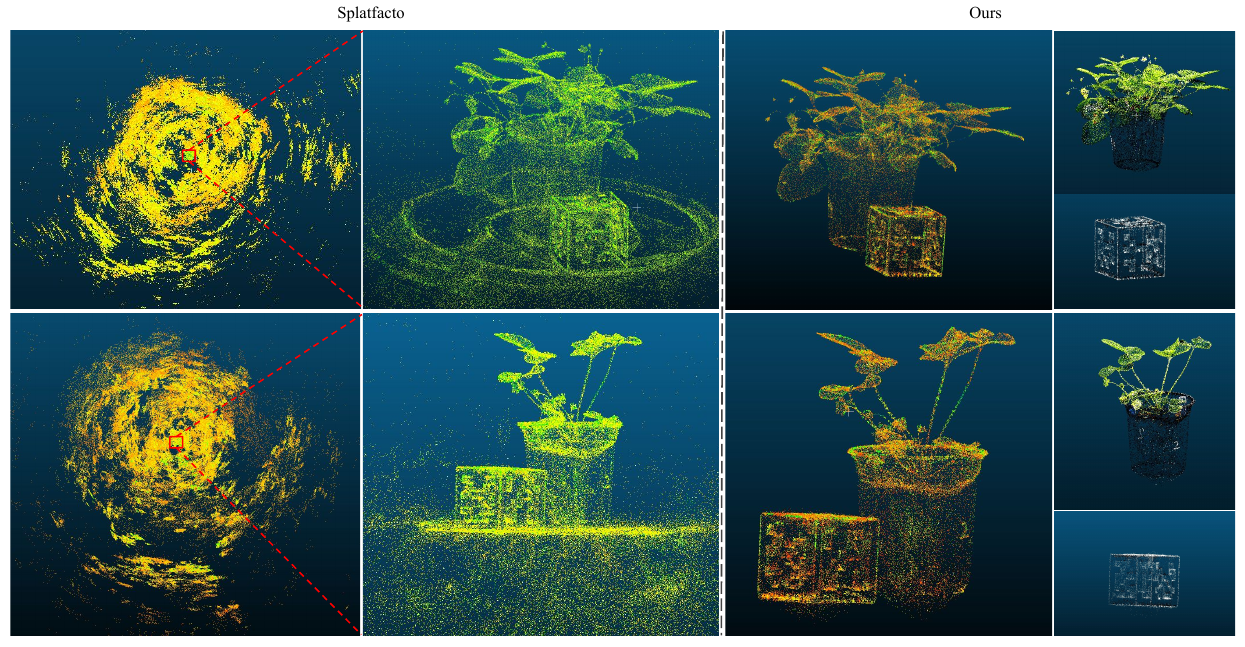}
  \vspace{1pt}
  \caption{Comparison of reconstructed point clouds between the baseline Splatfacto and the proposed object-centric 3DGS method. The left two columns show results from Splatfacto, which exhibit noisy and uneven point distributions due to background interference and redundant Gaussian primitives. The right columns display our method, which generates a cleaner, denser, and geometrically consistent point cloud with a clear separation between the plant and reference cube. The improved spatial organization and reduced background artifacts demonstrate the effectiveness of the integrated foreground masking and object-centric learning strategies.}
  \label{fig:pointcloud}
\end{figure*}

In addition to the quantitative comparisons, the qualitative point cloud visualizations in Figure~\ref{fig:pointcloud} further highlight the benefits of the proposed method. The Splatfacto baseline exhibits a high density of noisy Gaussian primitives scattered across the background, forming irregular clusters and circular artifacts that obscure plant geometry. This results in diffuse point distributions around the pot and reference cube, as seen in the left panels. In contrast, our object-centric 3DGS produces a well-organized and compact point cloud with a clear separation between the plant and its surroundings. The reconstructed plant canopy and reference cube are sharply defined, and the background noise is almost entirely eliminated. The dense yet uniform distribution of Gaussian centers demonstrates the effectiveness of mask-guided density control, which prunes redundant points while preserving fine geometric details in the leaves and fruits. These results qualitatively validate that integrating the background removal process directly into training yields cleaner, more accurate 3D structures that are crucial for downstream phenotyping and morphological analysis.

Overall, the combination of explicit Gaussian representation and foreground-guided optimization leads to reconstructions that are both computationally efficient and structurally faithful. These qualitative outcomes, together with the quantitative improvements in PSNR, SSIM, and LPIPS, confirm that the proposed method provides a promising foundation for high-fidelity, background-free plant modeling and phenotyping.

\subsection{Performance on plant traits}
To assess the applicability of the reconstructed 3D models for phenotypic analysis, we evaluated the accuracy of plant height and crown width estimation derived from the reconstructed point clouds. The object-centric 3DGS models were processed through the DBSCAN-PCA–based trait extraction pipeline, as described in Section~\ref{sec:oc_3dgs}. The reference cube in each scene provided the scale calibration, enabling conversion from reconstruction units to metric measurements (cm).

\begin{figure*}[!ht]
    \centering
    \subfloat[Height]{%
    \includegraphics[width=0.32\linewidth]{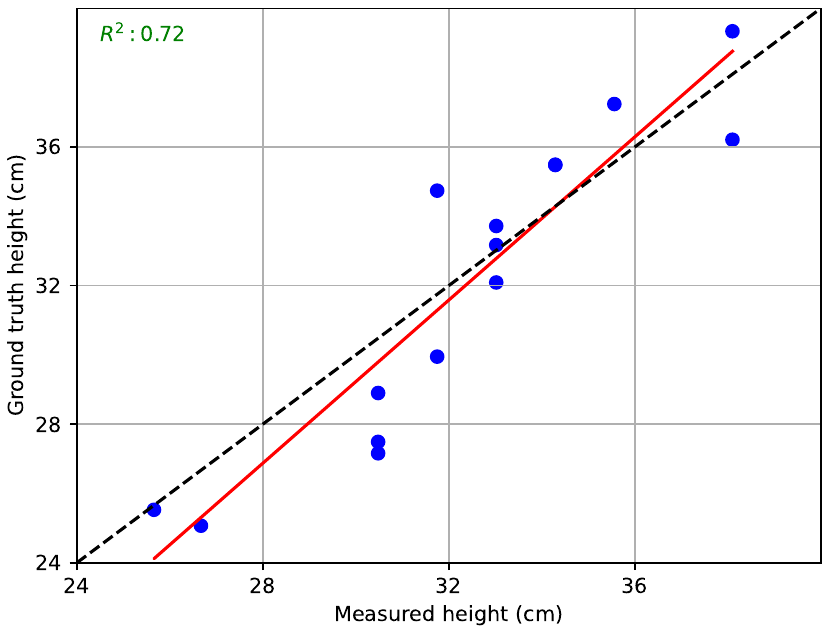}
        \label{fig:sub_blueberry}%
    }
    \vspace{-7pt}
    \subfloat[Width 1]{%
        \includegraphics[width=0.32\linewidth]{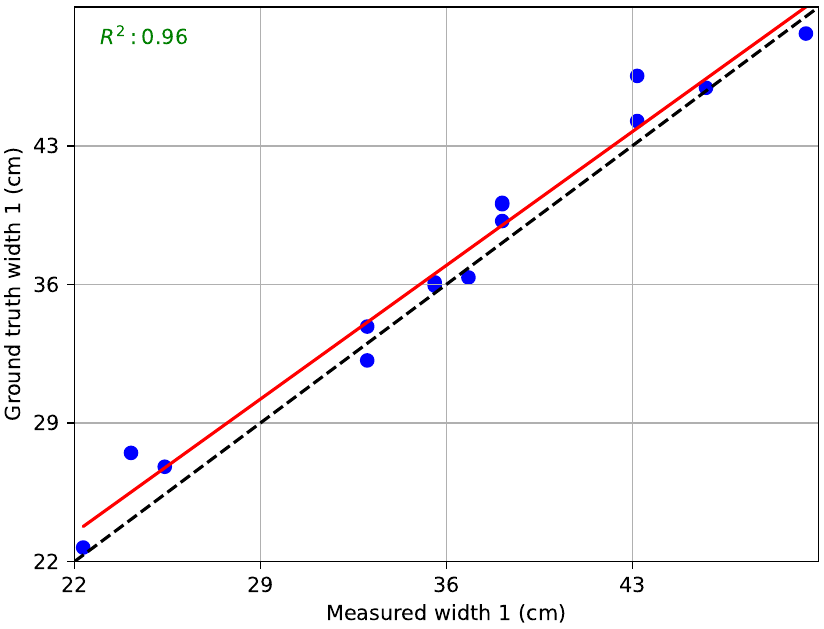}
        \label{fig:sub_strawberry}%
    } 
    \vspace{-7pt}
    \subfloat[Width 2]{%
        \includegraphics[width=0.32\linewidth]{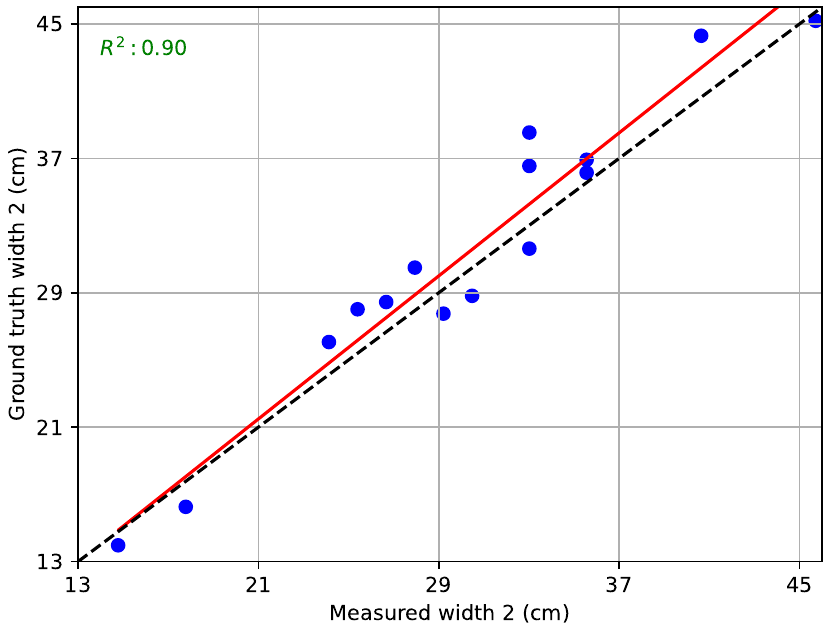}
        \label{fig: sub_apple}%
    }
    \vspace{10pt}
    \caption{Strawberry plant height and width estimation via DBSCAN and PCA. The red solid curve is the fitted line, and the black dashed curve is the ideal one.}
    \label{fig:plant_traits}
\end{figure*}

Figure~\ref{fig:plant_traits} presents the comparison between the estimated and ground-truth traits for 15 strawberry plants. Overall, the proposed approach achieved strong linear correlations across all traits, demonstrating that the reconstructed 3D geometry preserves accurate spatial information. Table~\ref{tab:plant_traits} summarizes the quantitative results. For plant height, the regression model yielded an $R^2$ of 0.72, with an RMSE of 1.81 cm, MAPE of 4.85\%, and an overall accuracy of 95.15\%. The estimation of Width 1 (major crown axis) achieved an $R^2$ of 0.96 and MAPE of 3.72\%, indicating high consistency with manual measurements. Similarly, Width 2 (minor crown axis) obtained an $R^2$  of 0.90, with an RMSE of 2.43 cm and accuracy of 92.94\%. These results demonstrate that the combination of object-centric 3DGS reconstruction, DBSCAN-based plant segmentation, and PCA-driven dimensional analysis provides an effective and robust framework for extracting plant morphological traits. The method accurately captures structural variability across individual plants while maintaining centimeter-level precision.
 
\begin{table}[h]
\renewcommand{\arraystretch}{1.6}
\centering
\caption{Evaluation metrics for strawberry plant height and canopy width estimation.}
\label{tab:plant_traits}
\resizebox{0.45 \textwidth}{!}{%
\begin{tabular}{c|ccccc}
\hline
\multicolumn{1}{l|}{} & \textbf{$R^2$}   & \textbf{RMSE} & \textbf{MAPE}   & \textbf{MAE}  & \textbf{Acc}     \\ \hline \hline
Height                 & 0.72 & 1.81 & 4.85\% & 1.56 & 95.15\% \\ 
Width 1                & 0.96 & 1.59 & 3.72\% & 1.25 & 96.28\% \\ 
Width 2                & 0.90 & 2.43 & 7.06\% & 2.06 & 92.94\% \\ \hline
\end{tabular}
}
\end{table}

\section{Limitations and Discussions}
The proposed object-centric 3DGS framework demonstrates the advantages of integrating segmentation-guided training within neural rendering pipelines. Unlike traditional NeRF models, which require extensive rendering iterations to resolve fine details, 3DGS directly models radiance through explicit Gaussian primitives, allowing efficient optimization. The addition of alpha-channel masking further enhances convergence speed by focusing learning on object-relevant regions. However, several limitations remain that offer promising avenues for further research and refinement.

\subsection{Environmental robustness and lighting conditions}
The current experiments were conducted under indoor, controlled illumination conditions, which effectively minimized shadowing, specular reflections, and motion-induced noise. While this setup ensures reproducibility and precise evaluation, it may not fully represent real-world agricultural environments, where natural light variation, fluctuating illumination spectra, and wind-induced plant motion can significantly affect image quality. These environmental factors may cause inconsistencies in photometric cues, camera pose estimation, and Gaussian density optimization. Future research should investigate adaptive illumination correction and temporal denoising strategies, such as photometric calibration, exposure compensation, and video-based multi-view stabilization, to maintain reconstruction consistency under uncontrolled outdoor conditions. Moreover, incorporating physics-based rendering priors, such as LiDAR-guided geometric constraints \citep{hess2025splatad, xiao2024liv}, into the 3DGS optimization process could further enhance robustness to dynamic lighting and shadow variations in natural field environments.

\subsection{Extension to complex phenotypic traits}
While the current DBSCAN–PCA–based trait estimation approach effectively quantifies key geometric attributes such as plant height and canopy width, it remains limited to low-level morphological features. These parameters, though fundamental, do not capture higher-order physiological indicators that are often more predictive of yield potential and plant health. For example, traits such as leaf inclination angle distribution, canopy volume, and fruit density require finer structural modeling and semantic understanding beyond simple geometric boundaries. Integrating semantic segmentation networks \citep{ravi2024sam} could enable trait-specific feature extraction directly from Gaussian primitives. Additionally, coupling the 3DGS outputs with multi-modal sensing modalities, such as hyperspectral, multispectral, or thermal imaging, could facilitate the simultaneous analysis of morphological and physiological states (e.g., photosynthetic activity, water stress, and disease onset), thereby bridging structural and functional plant phenotyping \citep{li2025survey}.

\subsection{Scalability to multi-plant and field-scale applications}
The scalability of the proposed framework to multi-plant or field-scale scenarios remains an important challenge. The current pipeline is optimized for individual-plant reconstruction, simplifying DBSCAN-based segmentation and scale calibration using a single reference cube. However, extending the framework to dense crop rows or multi-object scenes introduces challenges related to inter-plant occlusion, viewpoint overlap, and Gaussian density balancing. As the number of Gaussian primitives increases approximately quadratically with scene complexity, memory usage and optimization time may rise substantially. Addressing these issues may require the development of hierarchical or tile-based 3DGS strategies, in which localized Gaussian clusters are trained independently and later fused through spatial registration. Alternatively, incorporating distributed GPU architectures, cloud-based reconstruction, or multi-agent reconstruction pipelines could enable efficient large-scale modeling of crop canopies in outdoor environments \citep{yu2025aerial, ham2024dragon}. These advancements are critical for transitioning the framework from controlled single-plant experiments to high-throughput, field-level phenotyping systems capable of operating directly in commercial agricultural settings.

\section{Conclusion}
In this paper, we developed an object-centric 3DGS framework for high-fidelity strawberry plant reconstruction and quantitative phenotyping. Unlike traditional NeRF-based methods that are computationally intensive and sensitive to background noise, the proposed approach integrates foreground preprocessing, RGBA-based loss masking, opacity-guided Gaussian culling, and background randomization to focus learning on the plant’s structural geometry while suppressing irrelevant background regions. A systematic comparison with established NeRF variants (Nerfacto, Instant-NGP, and Mip-NeRF) demonstrated that our method achieves superior performance in terms of PSNR, SSIM, and LPIPS, while significantly reducing training time and memory consumption. The integration of SAM 2–based segmentation enables a foreground-aware reconstruction process, leading to clean and structurally consistent point clouds that facilitate accurate plant trait estimation. Using the reconstructed point clouds, we further demonstrated centimeter-level accuracy in measuring plant height and crown width via DBSCAN clustering and PCA-based geometric analysis. These results confirm that the proposed framework not only enhances reconstruction quality but also supports reliable, non-destructive, and automated plant phenotyping, which holds strong potential for agricultural monitoring, breeding, and yield estimation.

\section*{Authorship Contribution}
\textbf{Jiajia Li}: Conceptualization, Investigation, Software, Writing – original draft; 
\textbf{Keyi Zhu}: Conceptualization, Investigation, Writing – review; 
\textbf{Qianwen Zhang}: Resources, Writing – review; 
\textbf{Dong Chen}: Conceptualization, Investigation, Writing – review; 
\textbf{Qi Sun}: Conceptualization, Investigation, Writing – review; 
\textbf{Zhaojian Li}: Supervision, Conceptualization, Resources, Writing – review.

\section*{Acknowledgment}
The authors would like to thank Mr. Moeen Ul Islam at Mississippi State University for his assistance in printing 3D ArUco marker cubes.

\typeout{}
\bibliography{ref}

\begin{thebibliography}{46}
\providecommand{\natexlab}[1]{#1}
\providecommand{\url}[1]{\texttt{#1}}
\expandafter\ifx\csname urlstyle\endcsname\relax
  \providecommand{\doi}[1]{doi: #1}\else
  \providecommand{\doi}{doi: \begingroup \urlstyle{rm}\Url}\fi

\bibitem[Abdi and Williams(2010)]{abdi2010principal}
H.~Abdi and L.~J. Williams.
\newblock Principal component analysis.
\newblock \emph{Wiley interdisciplinary reviews: computational statistics}, 2\penalty0 (4):\penalty0 433--459, 2010.

\bibitem[Assarsson and Moller(2000)]{assarsson2000optimized}
U.~Assarsson and T.~Moller.
\newblock Optimized view frustum culling algorithms for bounding boxes.
\newblock \emph{Journal of graphics tools}, 5\penalty0 (1):\penalty0 9--22, 2000.

\bibitem[Bao et~al.(2025)Bao, Ding, Huo, Liu, Li, Li, Gao, and Luo]{bao20253d}
Y.~Bao, T.~Ding, J.~Huo, Y.~Liu, Y.~Li, W.~Li, Y.~Gao, and J.~Luo.
\newblock 3d gaussian splatting: Survey, technologies, challenges, and opportunities.
\newblock \emph{IEEE Transactions on Circuits and Systems for Video Technology}, 2025.

\bibitem[Barron et~al.(2021)Barron, Mildenhall, Tancik, Hedman, Martin-Brualla, and Srinivasan]{barron2021mip}
J.~T. Barron, B.~Mildenhall, M.~Tancik, P.~Hedman, R.~Martin-Brualla, and P.~P. Srinivasan.
\newblock Mip-nerf: A multiscale representation for anti-aliasing neural radiance fields.
\newblock In \emph{Proceedings of the IEEE/CVF international conference on computer vision}, pages 5855--5864, 2021.

\bibitem[Bochkovskiy et~al.(2020)Bochkovskiy, Wang, and Liao]{bochkovskiy2020yolov4}
A.~Bochkovskiy, C.-Y. Wang, and H.-Y.~M. Liao.
\newblock Yolov4: Optimal speed and accuracy of object detection.
\newblock \emph{arXiv preprint arXiv:2004.10934}, 2020.

\bibitem[Chen and Wang(2024)]{chen2024survey}
G.~Chen and W.~Wang.
\newblock A survey on 3d gaussian splatting.
\newblock \emph{arXiv preprint arXiv:2401.03890}, 2024.

\bibitem[Chen et~al.(2025)Chen, Jiao, Jin, Qin, Ning, Yang, and Zhan]{chen2025plant}
J.~Chen, Y.~Jiao, F.~Jin, X.~Qin, Y.~Ning, M.~Yang, and Y.~Zhan.
\newblock Plant sam gaussian reconstruction (psgr): A high-precision and accelerated strategy for plant 3d reconstruction.
\newblock \emph{Electronics}, 14\penalty0 (11):\penalty0 2291, 2025.

\bibitem[Choi et~al.(2024)Choi, Park, Park, and Lee]{choi2024nerf}
H.-B. Choi, J.-K. Park, S.~H. Park, and T.~S. Lee.
\newblock Nerf-based 3d reconstruction pipeline for acquisition and analysis of tomato crop morphology.
\newblock \emph{Frontiers in Plant Science}, 15:\penalty0 1439086, 2024.

\bibitem[Chopra et~al.(2024)Chopra, Cladera, Murali, and Kumar]{chopra2024agrinerf}
S.~Chopra, F.~Cladera, V.~Murali, and V.~Kumar.
\newblock Agrinerf: Neural radiance fields for agriculture in challenging lighting conditions.
\newblock \emph{arXiv preprint arXiv:2409.15487}, 2024.

\bibitem[Ester et~al.(1996)Ester, Kriegel, Sander, Xu, et~al.]{ester1996density}
M.~Ester, H.-P. Kriegel, J.~Sander, X.~Xu, et~al.
\newblock A density-based algorithm for discovering clusters in large spatial databases with noise.
\newblock In \emph{kdd}, volume~96, pages 226--231, 1996.

\bibitem[Fiorani and Schurr(2013)]{fiorani2013future}
F.~Fiorani and U.~Schurr.
\newblock Future scenarios for plant phenotyping.
\newblock \emph{Annual review of plant biology}, 64\penalty0 (1):\penalty0 267--291, 2013.

\bibitem[{Fresh Produce Association of the Americas}(2024)]{freshproduce2024strawberry}
{Fresh Produce Association of the Americas}.
\newblock U.s. strawberry market annual report 2024.
\newblock Technical report, Fresh Produce, 2024.
\newblock URL \url{https://www.freshproduce.com/siteassets/files/reports/global-trade/2024/strawberries_annual_market_report_2024.pdf}.
\newblock Accessed: 2025-10-01.

\bibitem[Gao et~al.(2022)Gao, Gao, He, Lu, Xu, and Li]{gao2022nerf}
K.~Gao, Y.~Gao, H.~He, D.~Lu, L.~Xu, and J.~Li.
\newblock Nerf: Neural radiance field in 3d vision, a comprehensive review.
\newblock \emph{arXiv preprint arXiv:2210.00379}, 2022.

\bibitem[Giampieri et~al.(2012)Giampieri, Tulipani, Alvarez-Suarez, Quiles, Mezzetti, and Battino]{giampieri2012strawberry}
F.~Giampieri, S.~Tulipani, J.~M. Alvarez-Suarez, J.~L. Quiles, B.~Mezzetti, and M.~Battino.
\newblock The strawberry: Composition, nutritional quality, and impact on human health.
\newblock \emph{Nutrition}, 28\penalty0 (1):\penalty0 9--19, 2012.

\bibitem[Ham et~al.(2024)Ham, Michalkiewicz, and Balakrishnan]{ham2024dragon}
Y.~Ham, M.~Michalkiewicz, and G.~Balakrishnan.
\newblock Dragon: Drone and ground gaussian splatting for 3d building reconstruction.
\newblock In \emph{2024 IEEE International Conference on Computational Photography (ICCP)}, pages 1--12. IEEE, 2024.

\bibitem[He et~al.(2017)He, Gkioxari, Doll{\'a}r, and Girshick]{he2017mask}
K.~He, G.~Gkioxari, P.~Doll{\'a}r, and R.~Girshick.
\newblock Mask r-cnn.
\newblock In \emph{Proceedings of the IEEE international conference on computer vision}, pages 2961--2969, 2017.

\bibitem[Hess et~al.(2025)Hess, Lindstr{\"o}m, Fatemi, Petersson, and Svensson]{hess2025splatad}
G.~Hess, C.~Lindstr{\"o}m, M.~Fatemi, C.~Petersson, and L.~Svensson.
\newblock Splatad: Real-time lidar and camera rendering with 3d gaussian splatting for autonomous driving.
\newblock In \emph{Proceedings of the Computer Vision and Pattern Recognition Conference}, pages 11982--11992, 2025.

\bibitem[Jain et~al.(2024)Jain, Mirzaei, and Gilitschenski]{jain2024gaussiancut}
U.~Jain, A.~Mirzaei, and I.~Gilitschenski.
\newblock Gaussiancut: Interactive segmentation via graph cut for 3d gaussian splatting.
\newblock \emph{Advances in Neural Information Processing Systems}, 37:\penalty0 89184--89212, 2024.

\bibitem[Jiang et~al.(2025)Jiang, Sun, Chee, Li, and Fu]{jiang2025cotton3dgaussians}
L.~Jiang, J.~Sun, P.~W. Chee, C.~Li, and L.~Fu.
\newblock Cotton3dgaussians: Multiview 3d gaussian splatting for boll mapping and plant architecture analysis.
\newblock \emph{Computers and Electronics in Agriculture}, 234:\penalty0 110293, 2025.

\bibitem[Jiang and Li(2020)]{jiang2020convolutional}
Y.~Jiang and C.~Li.
\newblock Convolutional neural networks for image-based high-throughput plant phenotyping: a review.
\newblock \emph{Plant Phenomics}, 2020.

\bibitem[Kerbl et~al.(2023)Kerbl, Kopanas, Leimk{\"u}hler, and Drettakis]{kerbl20233d}
B.~Kerbl, G.~Kopanas, T.~Leimk{\"u}hler, and G.~Drettakis.
\newblock 3d gaussian splatting for real-time radiance field rendering.
\newblock \emph{ACM Trans. Graph.}, 42\penalty0 (4):\penalty0 139--1, 2023.

\bibitem[Kouloumprouka~Zacharaki et~al.(2024)Kouloumprouka~Zacharaki, Monaghan, Bromley, and Vickers]{kouloumprouka2024opportunities}
A.~Kouloumprouka~Zacharaki, J.~M. Monaghan, J.~R. Bromley, and L.~H. Vickers.
\newblock Opportunities and challenges for strawberry cultivation in urban food production systems.
\newblock \emph{Plants, People, Planet}, 6\penalty0 (3):\penalty0 611--621, 2024.

\bibitem[Lassner and Zollhofer(2021)]{lassner2021pulsar}
C.~Lassner and M.~Zollhofer.
\newblock Pulsar: Efficient sphere-based neural rendering.
\newblock In \emph{Proceedings of the IEEE/CVF Conference on Computer Vision and Pattern Recognition}, pages 1440--1449, 2021.

\bibitem[Li et~al.(2025)Li, Qi, Nabaei, Liu, Chen, Zhang, Yin, and Li]{li2025survey}
J.~Li, X.~Qi, S.~H. Nabaei, M.~Liu, D.~Chen, X.~Zhang, X.~Yin, and Z.~Li.
\newblock A survey on 3d reconstruction techniques in plant phenotyping: from classical methods to neural radiance fields (nerf), 3d gaussian splatting (3dgs), and beyond.
\newblock \emph{arXiv preprint arXiv:2505.00737}, 2025.

\bibitem[Li et~al.(2014)Li, Zhang, and Huang]{li2014review}
L.~Li, Q.~Zhang, and D.~Huang.
\newblock A review of imaging techniques for plant phenotyping.
\newblock \emph{Sensors}, 14\penalty0 (11):\penalty0 20078--20111, 2014.

\bibitem[Liu et~al.(2023)Liu, Liang, and Kang]{liu2023molecular}
Z.~Liu, T.~Liang, and C.~Kang.
\newblock Molecular bases of strawberry fruit quality traits: Advances, challenges, and opportunities.
\newblock \emph{Plant Physiology}, 193\penalty0 (2):\penalty0 900--914, 2023.

\bibitem[Ma{\'c}kiewicz and Ratajczak(1993)]{mackiewicz1993principal}
A.~Ma{\'c}kiewicz and W.~Ratajczak.
\newblock Principal components analysis (pca).
\newblock \emph{Computers \& Geosciences}, 19\penalty0 (3):\penalty0 303--342, 1993.

\bibitem[Markin et~al.(2024)Markin, Pryadilshchikov, Komarichev, Rakhimov, Wonka, and Burnaev]{markin2024t}
A.~Markin, V.~Pryadilshchikov, A.~Komarichev, R.~Rakhimov, P.~Wonka, and E.~Burnaev.
\newblock T-3dgs: Removing transient objects for 3d scene reconstruction.
\newblock \emph{arXiv preprint arXiv:2412.00155}, 2024.

\bibitem[Mildenhall et~al.(2021)Mildenhall, Srinivasan, Tancik, Barron, Ramamoorthi, and Ng]{mildenhall2021nerf}
B.~Mildenhall, P.~P. Srinivasan, M.~Tancik, J.~T. Barron, R.~Ramamoorthi, and R.~Ng.
\newblock Nerf: Representing scenes as neural radiance fields for view synthesis.
\newblock \emph{Communications of the ACM}, 65\penalty0 (1):\penalty0 99--106, 2021.

\bibitem[M{\"u}ller et~al.(2022)M{\"u}ller, Evans, Schied, and Keller]{muller2022instant}
T.~M{\"u}ller, A.~Evans, C.~Schied, and A.~Keller.
\newblock Instant neural graphics primitives with a multiresolution hash encoding.
\newblock \emph{ACM transactions on graphics (TOG)}, 41\penalty0 (4):\penalty0 1--15, 2022.

\bibitem[Ndikumana et~al.(2024)Ndikumana, Lee, Yoo, Yeboah, Park, Lee, Yeoung, and Kim]{ndikumana2024development}
J.~N. Ndikumana, U.~Lee, J.~H. Yoo, S.~Yeboah, S.~H. Park, T.~S. Lee, Y.~R. Yeoung, and H.~S. Kim.
\newblock Development of a deep-learning phenotyping tool for analyzing image-based strawberry phenotypes.
\newblock \emph{Frontiers in Plant Science}, 15:\penalty0 1418383, 2024.

\bibitem[Ravi et~al.(2024)Ravi, Gabeur, Hu, Hu, Ryali, Ma, Khedr, R{\"a}dle, Rolland, Gustafson, et~al.]{ravi2024sam}
N.~Ravi, V.~Gabeur, Y.-T. Hu, R.~Hu, C.~Ryali, T.~Ma, H.~Khedr, R.~R{\"a}dle, C.~Rolland, L.~Gustafson, et~al.
\newblock Sam 2: Segment anything in images and videos.
\newblock \emph{arXiv preprint arXiv:2408.00714}, 2024.

\bibitem[Rogge and Stricker(2025)]{rogge2025object}
M.~Rogge and D.~Stricker.
\newblock Object-centric 2d gaussian splatting: Background removal and occlusion-aware pruning for compact object models.
\newblock \emph{arXiv preprint arXiv:2501.08174}, 2025.

\bibitem[Ronneberger et~al.(2015)Ronneberger, Fischer, and Brox]{ronneberger2015u}
O.~Ronneberger, P.~Fischer, and T.~Brox.
\newblock U-net: Convolutional networks for biomedical image segmentation.
\newblock In \emph{International Conference on Medical image computing and computer-assisted intervention}, pages 234--241. Springer, 2015.

\bibitem[Rota~Bul{\`o} et~al.(2024)Rota~Bul{\`o}, Porzi, and Kontschieder]{rota2024revising}
S.~Rota~Bul{\`o}, L.~Porzi, and P.~Kontschieder.
\newblock Revising densification in gaussian splatting.
\newblock In \emph{European Conference on Computer Vision}, pages 347--362. Springer, 2024.

\bibitem[Shen et~al.(2025)Shen, Jing, Deng, Jia, and Wu]{shen2025plantgaussian}
P.~Shen, X.~Jing, W.~Deng, H.~Jia, and T.~Wu.
\newblock Plantgaussian: exploring 3d gaussian splatting for cross-time, cross-scene, and realistic 3d plant visualization and beyond.
\newblock \emph{The Crop Journal}, 13\penalty0 (2):\penalty0 607--618, 2025.

\bibitem[Tancik et~al.(2023)Tancik, Weber, Ng, Li, Yi, Kerr, Wang, Kristoffersen, Austin, Salahi, Ahuja, McAllister, and Kanazawa]{nerfstudio}
M.~Tancik, E.~Weber, E.~Ng, R.~Li, B.~Yi, J.~Kerr, T.~Wang, A.~Kristoffersen, J.~Austin, K.~Salahi, A.~Ahuja, D.~McAllister, and A.~Kanazawa.
\newblock Nerfstudio: A modular framework for neural radiance field development.
\newblock In \emph{ACM SIGGRAPH 2023 Conference Proceedings}, SIGGRAPH '23, 2023.

\bibitem[Tulipani et~al.(2011)Tulipani, Marzban, Herndl, Laimer, Mezzetti, and Battino]{tulipani2011influence}
S.~Tulipani, G.~Marzban, A.~Herndl, M.~Laimer, B.~Mezzetti, and M.~Battino.
\newblock Influence of environmental and genetic factors on health-related compounds in strawberry.
\newblock \emph{Food Chemistry}, 124\penalty0 (3):\penalty0 906--913, 2011.

\bibitem[Xiao et~al.(2024)Xiao, Liu, Chen, and Hu]{xiao2024liv}
R.~Xiao, W.~Liu, Y.~Chen, and L.~Hu.
\newblock Liv-gs: Lidar-vision integration for 3d gaussian splatting slam in outdoor environments.
\newblock \emph{IEEE Robotics and Automation Letters}, 2024.

\bibitem[Yang et~al.(2024)Yang, Lu, Xie, Guo, Fang, Fu, Hu, Sun, and Cen]{yang2024paniclenerf}
X.~Yang, X.~Lu, P.~Xie, Z.~Guo, H.~Fang, H.~Fu, X.~Hu, Z.~Sun, and H.~Cen.
\newblock Paniclenerf: low-cost, high-precision in-field phenotyping of rice panicles with smartphone.
\newblock \emph{Plant Phenomics}, 6:\penalty0 0279, 2024.

\bibitem[Yu et~al.(2025)Yu, Wang, Jiang, Zhang, Zhang, and Li]{yu2025aerial}
J.~Yu, H.~Wang, S.~Jiang, X.~Zhang, D.~Zhang, and Q.~Li.
\newblock Aerial-ground image feature matching via 3d gaussian splatting-based intermediate view rendering.
\newblock \emph{arXiv preprint arXiv:2509.19898}, 2025.

\bibitem[Zhang et~al.(2024)Zhang, Wang, Ni, Dong, Tang, Sun, and Wang]{zhang2024neural}
J.~Zhang, X.~Wang, X.~Ni, F.~Dong, L.~Tang, J.~Sun, and Y.~Wang.
\newblock Neural radiance fields for multi-scale constraint-free 3d reconstruction and rendering in orchard scenes.
\newblock \emph{Computers and Electronics in Agriculture}, 217:\penalty0 108629, 2024.

\bibitem[Zhao et~al.(2024)Zhao, Ying, Pan, Yi, Chen, Hu, and Kang]{zhao2024exploring}
J.~Zhao, W.~Ying, Y.~Pan, Z.~Yi, C.~Chen, K.~Hu, and H.~Kang.
\newblock Exploring accurate 3d phenotyping in greenhouse through neural radiance fields.
\newblock \emph{arXiv preprint arXiv:2403.15981}, 2024.

\bibitem[Zheng et~al.(2022)Zheng, Abd-Elrahman, Whitaker, and Dalid]{zheng2022deep}
C.~Zheng, A.~Abd-Elrahman, V.~M. Whitaker, and C.~Dalid.
\newblock Deep learning for strawberry canopy delineation and biomass prediction from high-resolution images.
\newblock \emph{Plant Phenomics}, 2022.

\bibitem[Zheng et~al.(2024)Zheng, Zhou, Shao, Liu, Zhang, Nie, and Liu]{zheng2024gps}
S.~Zheng, B.~Zhou, R.~Shao, B.~Liu, S.~Zhang, L.~Nie, and Y.~Liu.
\newblock Gps-gaussian: Generalizable pixel-wise 3d gaussian splatting for real-time human novel view synthesis.
\newblock In \emph{Proceedings of the IEEE/CVF conference on computer vision and pattern recognition}, pages 19680--19690, 2024.

\bibitem[Zhu et~al.(2024)Zhu, Huang, and Li]{zhu2024three}
X.~Zhu, Z.~Huang, and B.~Li.
\newblock Three-dimensional phenotyping pipeline of potted plants based on neural radiation fields and path segmentation.
\newblock \emph{Plants}, 13\penalty0 (23):\penalty0 3368, 2024.

\end{thebibliography}
\end{document}